%% file: main.tex
\documentclass{article}
\usepackage{arxiv}
\usepackage[utf8]{inputenc}

\usepackage{ amsmath , amssymb , amsthm }

\title{Unsupervised Energy-based Out-of-distribution Detection using Stiefel-Restricted Kernel Machine}

\usepackage{mathtools}
\mathtoolsset{showonlyrefs=true}

\usepackage{makecell}
\usepackage{booktabs}
\usepackage{lipsum}
\usepackage{url}
\usepackage{multicol}
\usepackage{multirow}
\usepackage{placeins}
\usepackage{tikz}
\usepackage{pgfplots}
\usetikzlibrary{shapes,arrows,backgrounds,positioning}
\usetikzlibrary{calc}
\tikzstyle{block} = [draw,rectangle,thick,minimum height=2em,minimum width=2em]
\tikzstyle{neuron}=[draw,circle,minimum size=12pt,inner sep=0pt, fill=white]
\tikzstyle{box}=[draw, fill=black!0, rounded corners]

\DeclareMathOperator{\Tr}{Tr}
\DeclareMathOperator{\St}{St}
\DeclareMathOperator{\diag}{\mathrm{diag}} 
\newcommand{\Dtrainin}{\mathcal{D}^{\text{train}}_{\text{in}}}
\newcommand{\Dtestout}{\mathcal{D}^{\text{test}}_{\text{out}}}
\newcommand{\bm}[1]{#1} %
\newcommand{\h}[2]{\|\bm{h}_{#1}^{#2}\|_2^2} %
\newcommand{\phimap}[2]{\|\bm{\phi}_{\bm{#1}}(\bm{x}_{#2})\|_2^2} %
\newcommand{\corr}[4]{2 \bm{\phi}_{\bm{#1}}^\top(\bm{x}_{#2}) U^{#3} \bm{h}_{#4}} %
\newcommand{\loss}[4]{L_{\bm{#1},U^{#2}}(\bm{x}_{#3},\bm{\phi}_{\bm{#4}}(\bm{x}_{#3}))} %

\usepackage{etoolbox}
\makeatletter
\patchcmd{\@makecaption}
  {\scshape}
  {}
  {}
  {}
\makeatletter
\patchcmd{\@makecaption}
  {\\}
  {.\ }
  {}
  {}
\makeatother

\begin{document}

\author{
 Francesco Tonin \\
  Department of Electrical Engineering\\
  ESAT-STADIUS, KU Leuven\\
  Kasteelpark Arenberg 10, B-3001 Leuven, Belgium \\
  \texttt{francesco.tonin@esat.kuleuven.be} \\  %
   \And
 Arun Pandey \\
  Department of Electrical Engineering\\
  ESAT-STADIUS, KU Leuven\\
  Kasteelpark Arenberg 10, B-3001 Leuven, Belgium \\
  \texttt{arun.pandey@esat.kuleuven.be} \\  %
  \And
 Panagiotis Patrinos \\
  Department of Electrical Engineering\\
  ESAT-STADIUS, KU Leuven\\
  Kasteelpark Arenberg 10, B-3001 Leuven, Belgium \\
  \texttt{panos.patrinos@esat.kuleuven.be} \\  %
  \And
   Johan A. K. Suykens \\
  Department of Electrical Engineering\\
  ESAT-STADIUS, KU Leuven\\
  Kasteelpark Arenberg 10, B-3001 Leuven, Belgium \\
  \texttt{johan.suykens@esat.kuleuven.be} \\  %
}

\maketitle

\begin{abstract}
Detecting out-of-distribution (OOD) samples is an essential requirement for the deployment of machine learning systems in the real world.
Until now, research on energy-based OOD detectors has focused on the softmax confidence score from a pre-trained neural network classifier with access to class labels.
In contrast, we propose an unsupervised energy-based OOD detector leveraging the Stiefel-Restricted Kernel Machine (St-RKM). Training requires minimizing an objective function with an autoencoder loss term and the RKM energy where the interconnection matrix lies on the Stiefel manifold. 
Further, we 
outline multiple energy function definitions based on the RKM framework and discuss their utility.
In the experiments on standard datasets, the proposed method improves over the existing energy-based OOD detectors and deep generative models.
Through several ablation studies,
we further illustrate the merit of each proposed energy function on the OOD detection performance.
\end{abstract}

\section{Introduction}
While modern Deep Learning classifiers can achieve exceptional accuracy in many domains, deploying such models in an open-world setting is not trivial. This requires algorithms to detect anomalous test samples that do not belong to the data distribution on which the model was trained on.
In fact, it has been previously observed that neural network classifiers can provide predictions with high confidence on adversarial examples~\cite{goodfellow2015}, on distributions far from the training distribution \cite{hendrycks2017} and even on Gaussian noise \cite{hendrycks2017}.
In contrast, for safety-critical applications, one would like their machine learning systems to flag potentially anomalous test samples so that erroneous predictions are prevented and human intervention can take place for further assessment.

For the out-of-distribution (OOD) detection task, we aim to build a binary classifier that, given a test sample $x$, decides whether $x$ belongs to the training distribution or not. Formally, consider a training dataset $\mathcal{D}^{\text{train}}_{\text{in}}$ drawn i.i.d. from a data distribution $P$, called the in-distribution. Let $Q$ be an unknown distribution, called the out-distribution, from which anomalous examples are drawn. The out-of-distribution detection task then involves computing an anomaly score $s(x) \in \mathbb{R}$, where $x \in \mathbb{R}^D$ is a test sample. The convention we use throughout the text is that the higher the anomaly score, the more likely it is that $x$ is sampled from $Q$. Note that the OOD detection differs from the related problem of anomaly detection because the latter assumes that the outliers are present in the training set. On the contrary, the dataset is assumed to be not contaminated in the former case. %
Several approaches to anomaly detection exist, such as in the domain of clustering and robust statistics (for a comprehensive review, see \cite{rousseeuw2011}, \cite{hodge2004}). In anomaly detection, the models are built to detect and penalize the outliers present in the training data. For instance, this was studied in the context of robust Restricted Kernel Machines (RKMs) \cite{robustrkm}. However, in such models an important hyperparameter to consider before training is the contamination rate, i.e., the percentage of anomalies in the training data.

In this paper, we investigate OOD detection with an energy function based on the Stiefel-Restricted Kernel Machine (St-RKM) framework \cite{strkm}. In this energy-based framework, the model parameters are learned in an unsupervised manner via manifold optimization where the interconnection matrix lies on the Stiefel manifold. We propose multiple energy function definitions based on the St-RKM objective, provide insights into their mathematical meaning, and   discuss their practical implications for the OOD detection performance. 
We show the effectiveness of the proposed method in computer vision tasks and for time series data.

While the state of the art in energy-based OOD detection, proposed by Liu et al. \cite{liu2020a}, relies on a neural classifier trained in a supervised manner on $\Dtrainin$, the training phase of our method is unsupervised. Further, in \cite{liu2020a} a fine-tuning step employing an auxiliary dataset sampled from $Q$ is needed to achieve significantly improved OOD detection performance. On the contrary, the training phase of our method is agnostic to the OOD detection task, and therefore only needs in-distribution samples. This  is particularly significant when anomalous examples are available in limited quantity or are expensive to collect. In other words, the proposed model is insensitive to the imbalance of normal/abnormal samples in the dataset.

\begin{figure*}[t]
    \centering
    \resizebox{0.8\textwidth}{!}{%
    \begin{tikzpicture}
\node[block, label=below:2. Energy Scores Computation] (1) {$E_{\text{energy}}(x) =  \| (\mathbb{I} - \bm{U}\bm{U}^\top) \bm{\phi}_{\bm{\theta}}(\bm{x})\|_2^2 + \lambda  L_{\bm{\xi},U}(\bm{x},\bm{\phi}_{\bm{\theta}}(\bm{x})) 
$};

\node [draw, cylinder, shape border rotate=90, aspect=0.45, %
minimum height=40, minimum width=40, left=of 1, xshift=-5cm, yshift=-1.5cm] (cyc) {$D_{in}^{train}$};

\node[right=of cyc] (x) {$\mathcal{X}$};

\node (b1) [opacity=1, box, text width=0.35cm, text height=2cm, right=of x, label=below:$\mathcal{F}$] {};
\node (b2) [opacity=1, box, text width=0.35cm, text height=2cm, right=of b1, label=below:$\mathcal{H}$] {};

\node (v1)[neuron, right=of x, yshift=0.8cm, xshift=2pt] {};
\node (v2)[neuron, below of = v1, yshift=6pt] {};
\node (v3)[neuron, below of = v2, yshift=6pt] {};

\node (v4)[neuron, right=of b1, yshift=0.8cm, xshift=2pt] {};
\node (v5)[neuron, below of = v4, yshift=6pt] {};
\node (v6)[neuron, below of = v5, yshift=6pt] {};

\draw[ >=stealth,->] (cyc) -- (x) node [midway,above] {};
\draw[ >=stealth,->] (x) -- (b1.120) node [midway,above] {$\phi^{\top}$};
\draw[ >=stealth,<-] (x) -- (b1.240) node [midway,below] {$\psi^{\top}$};

\draw[ >=stealth,->] (b1.60) -- (b2.120) node [midway,above] {$U^{\top}$};
\draw[ >=stealth,<-] (b1.300) -- (b2.240) node [midway,below] {$U$};

\draw[block] ($(cyc.south west) - (12pt,40pt)$)
rectangle ($(b2.north east) + (8pt,22pt)$) node [midway, below, yshift=-1.8cm] {1. St-RKM Model Training} ;

\draw[ >=stealth,->, thick, xshift=1.2cm] (-6,0) -- (-5.3,0); %

\node[below left=of 1, yshift=-0.5cm, xshift=2cm] (xs) {$x_{i}$};
\node[below=of xs, yshift=1.1cm, xshift=0.5cm] (xsi) {$i\in\{0, \cdots , 3\}$};

\draw[ >=stealth,->] (-2.8,-2.1) -- (-2.0,-2.1);
\draw[ >=stealth,->] (1.0,-2.1) -- (1.4,-2.1);

\node[xshift=4cm] (n1) {};
\node[xshift=4cm, yshift=-2.1cm] (n2) {};

\draw[>=stealth,->, thick]([xshift=-0.05cm,yshift=0cm]n1.east) to[out=0, in=0, looseness=1.1] ([xshift=-0.2cm,yshift=0cm]n2.east);

\begin{axis}[ at={($(xs)+(2.1cm,0cm)$)}, anchor=west, scale=0.3]
\addplot [dashed,color=blue]
plot coordinates {
  (0,0.5)
  (0.5,0.5)
  (3,0.5)
};

\node[] at (axis cs: 2.5,.3) {$\gamma$};
\node[] at (axis cs: 0.8, 1.3) {$E(x)$};

\addplot[smooth,color=black, mark=*]
plot coordinates {
	(0,0)
	(1,1)
	(2,0.3)
	(2.5,1.5)
};
\end{axis}

\node[block, thin, right=of xs, xshift=3.5cm,text width=1.8cm, align=center] (ood) {Flag $x_1$, $x_3$ as OOD points};

\draw[block] ($(xs.south west) - (12pt,30pt)$)
rectangle ($(ood.north east) + (6pt,12pt)$) node [midway, below, yshift=-1.1cm] (3) {3. Input-level validation in Deployment};

\end{tikzpicture}
}%
    \caption{Schematic figure illustrating the pipeline of training and detection of Out-of-Distribution (OOD) points. First, the model is trained on the in-distribution dataset with the full energy (objective) function, where $\mathcal{F}$ is the feature space and $\mathcal{H}$ is the latent subspace. Then, the threshold $\gamma$ is selected such that, with the chosen energy metric, the scores of 95\% of training points are below the threshold value. Lastly, in the evaluation phase, a test sample $x_i$ is passed through the model and its energy score is calculated with the desired metric. If the score is below/above the threshold, the test point is flagged as in/out-of-distribution sample.}
    \label{fig:schematic}
\end{figure*}
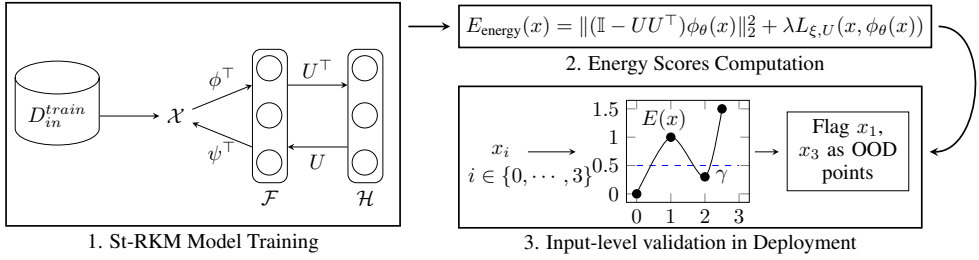

\textbf{Contributions.} The main contributions are as follows:
\begin{itemize}
\item We propose an energy-based OOD detection method leveraging the St-RKM  framework where the training procedure is agnostic to the OOD task. Contrary to the previous approaches using a pre-trained neural network, here the training is unsupervised.
\item We propose multiple energy function definitions for OOD detection based on the St-RKM objective, discussing their properties and scenarios in which they may perform best in practice. 
\item We empirically evaluate our method on multiple performance metrics, including the false-positive rate at 95\% true positive rate (FPR95), area under the receiver operating characteristic curve (ROC), and area under the precision-recall curve (PR), showing that it improves over generative models and state-of-the-art energy-based OOD detection methods using a pre-trained neural network classifier. We also provide further discussion on the multiple energy function definitions by analyzing their resulting energy distributions.
\end{itemize}

\section{Related Work}
Though many classical statistical models such as kernel density estimators, mixture models, and Principal Component Analysis (PCA) \cite{pca} have been studied extensively in the past for OOD detection (for a review, see \cite{pimentel2014}), this section reviews closely related approaches based on neural networks. 
\subsection{OOD detection using pre-trained neural networks}
In the context of neural network classifiers, Hendrycks et al. \cite{hendrycks2017} show that the prediction probability of out-of-distribution samples is usually lower than the prediction probability of in-distribution samples. Building upon this observation, \cite{hendrycks2017} proposes an algorithm based on the maximum predicted probability from the softmax distribution given by the output layer of a neural network. Formally, consider a pre-trained neural network classifier $f(x): \mathbb{R}^D \to \mathbb{R}^K$ that maps an input $x$ to $K$ real-valued outputs, also known as logits. The score function proposed by \cite{hendrycks2017} is:
\begin{equation} \label{eq:hendrycks2017}
    s(x) = - \max_j \frac{e^{{f_j(x)}}}{\sum_{i=1}^K e^{f_i(x)}},
\end{equation}
where $f_j(x)$ is the $j^{\text{th}}$ component of $f$.
By further developing this approach, Liang et al. \cite{liang2018} introduce small perturbations to the test input $x$ and \eqref{eq:hendrycks2017} is augmented by temperature scaling, The score proposed by \cite{liang2018} is then: %
\begin{equation} \label{eq:liang2018}
    s(x) = - \max_j \frac{e^{f_j(\tilde{x})/T}}{\sum_{i=1}^K e^{f_i(\tilde{x})/T}},
\end{equation}
where $\tilde{x}$ is the preprocessed input according to Eq. (2) in \cite{liang2018} and $T \in \mathbb{R}_{>0}$ is a temperature scaling parameter. These two additions are shown to increase the gap in scores between OOD and in-distribution samples.

Later, Lee et al. \cite{lee2018} propose a Mahalanobis distance-based anomaly score. First, they define $K$ class-conditional Gaussian distributions as $P(f(x)|y=k)  = \mathcal{N}(f(x)|\mu_k,\Sigma)$ for all $k=1,2,\dots,K$, whose parameters are estimated from $\mathcal{D}^{\text{train}}_{\text{in}}$. The score function is then defined using the Mahalanobis distance between the closest class-conditional Gaussian distribution and the test sample $x$ and is given by
\begin{equation} \label{eq:lee2018}
    s(x) = \min_k (f(\tilde{x})-\mu_k)^T \Sigma^{-1} (f(\tilde{x})-\mu_k),
\end{equation}
where $\tilde{x}$ is the preprocessed input according to Eq. (4) in \cite{lee2018}.

Contrary to the methods reviewed in this subsection, our proposed method is not based on a pre-trained neural network classifier. In fact, our method is based on an encoder-decoder architecture and is completely unsupervised; furthermore, its training phase is not specific to the OOD detection task.

	\begin{figure*}
		\centering
		\def\svgwidth{0.8\linewidth}
		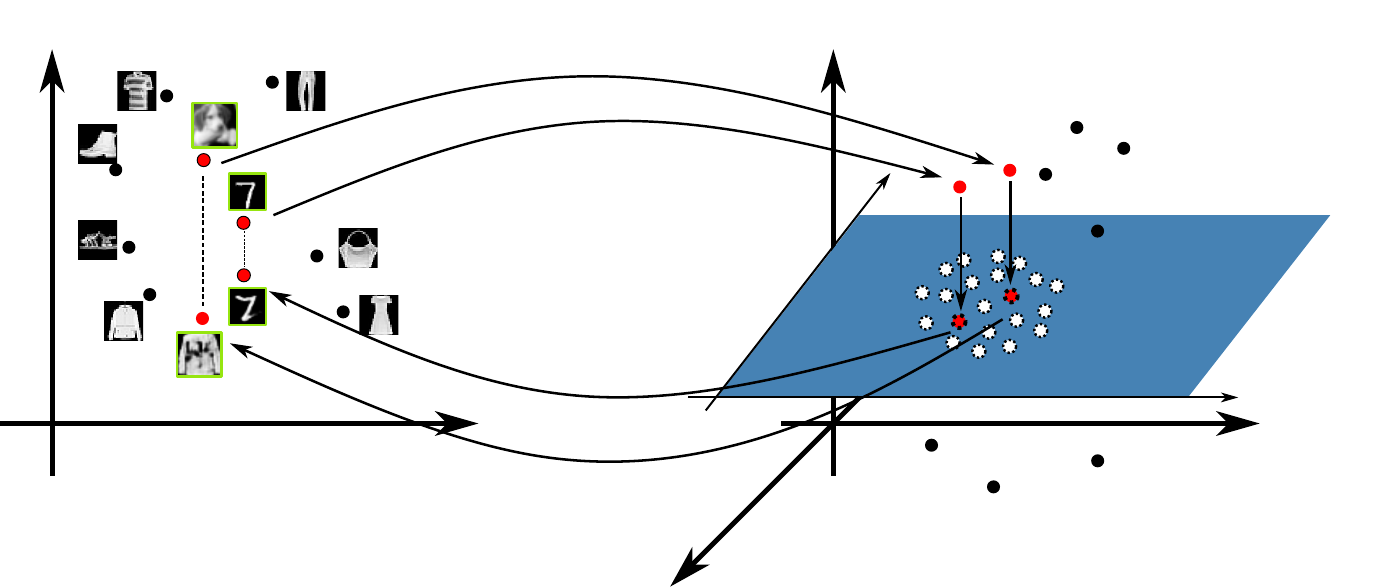
		\caption{Illustration of the effect of the St-RKM objective function when $\Dtrainin$ = Fashion-MNIST and $\Dtestout$ = \{CIFAR-10, MNIST\}. For datasets whose distribution is closer to Fashion-MNIST, the AutoEncoder error is smaller (norm of the dashed line). Hence, the KPCA reconstruction error (norm of the solid line in latent space) becomes a more relevant metric for detecting OOD samples. For datasets whose distribution is dissimilar such as CIFAR-10, both the errors are significant and, hence, the $E_{\text{FullEnergy}}$ becomes more relevant to flag OOD samples.}
		\label{fig:ood_cifar}
	\end{figure*}

\subsection{OOD detection with fine-tuning}
Differently from the methods above, approaches requiring additional training on top of the pre-trained neural network $f$ have been proposed. In these methods, the neural network's weights are updated such that the discriminative power of its outputs is boosted with respect to out-of-distribution detection. In the follow-up paper, Hendrycks et al. \cite{hendrycks2019} propose a method called ``Outlier Exposure" (OE), which fine-tunes the pre-trained classifier $f$ using an auxiliary dataset $\mathcal{D}^{\text{OE}}_{\text{out}}$ drawn from $Q$. In this way, the score function remains \eqref{eq:hendrycks2017}, but the neural network's weights are first optimized by minimizing
\begin{equation} \label{eq:hendrycks2019}
    \mathbb{E}_{(x,y)\sim\mathcal{D}^{\text{train}}_{\text{in}}} \left[ \mathcal{L}(f(x),y) + \lambda \mathbb{E}_{x'\sim\mathcal{D}^{\text{OE}}_{\text{out}}} \left[ \mathcal{L}_{\text{OE}}(f(x'), f(x), y) \right] \right],
\end{equation}
where $x'$ is a sample from the auxiliary OOD dataset, while $x$ is a sample from $\Dtrainin$. $\mathcal{L}_{OE}$ is a design choice depending on the kind of task. For instance, \cite{hendrycks2019} proposes to use the cross-entropy loss when $\mathcal{L}$ is based on the maximum predicted probability from the softmax distribution.

The training phase of the proposed method is agnostic to the OOD detection task and does not require an ad-hoc fine-tuning step.

\subsection{OOD detection using deep generative models}
Deep generative models approximate the true training data distribution $P$ with a density $p(x;\theta)$, where $\theta$ denotes the parameters of the employed deep neural networks. In this framework, test points that lie in the low-density regions can be labeled as out-of-distribution. To compare deep generative models with classical methods, {\v S}kv{\'a}ra et al. \cite{skvara2018} conducted experiments in OOD detection with the variational autoencoder (VAE) \cite{vae} and the generative adversarial network (GAN) \cite{gan} in several domains, concluding that deep generative models can outperform conventional methods if their hyperparameters are well tuned. For VAEs, they use the reconstruction error as the score function. For GANs, they propose the following score:
\begin{equation} \label{eq:gan}
    s(x) = - (1-\lambda) \log (d_\theta(x)) + \lambda \|x-g_\phi(z)\|_2,
\end{equation}
where $d_\theta$ is the discriminator, $g_\phi$ is the generator, $\lambda$ is a scaling parameter and $z \sim p(z)$, where $p(z)$ is the normal distribution. In the experimental evaluation of \cite{skvara2018}, $\lambda = 0$ led to the best AUROC performance; hence we use this choice in Section \ref{exp}.
More recently, Nalisnick et al. \cite{nalisnick2019} show that the likelihood from deep generative models such as VAEs cannot effectively separate in and out-of-distribution samples, as the latter can be assigned higher likelihood than the former. However, this result is based on the likelihood computed from deep generative models, and it is possible that the performance of these models can be improved using different score functions.

Our method, based on the St-RKM \cite{strkm}, is also a generative model; however, instead of the likelihood, we take the energy function derived from RKMs as the score.

\subsection{OOD detection using energy-based models}
Energy-based models \cite{lecun2006} employ an energy function $E(\cdot): \mathbb{R}^D \to \mathbb{R}$ that associates a scalar value to an input $x$ such that more likely inputs are associated with lower energies. The energy can therefore be used as a score function: inputs with lower energy scores are labeled as in-distribution and inputs with higher energy scores are labeled as out-of-distribution.

In this context, Grathwohl et al. \cite{grathwohl2020} observe that, given a neural network classifier $f$ with parameters $\theta$, the density $p(x;\theta)$ can be written using the logits of $f$ by marginalizing over the $K$ classes:
\begin{equation}
    p(x;\theta) = \sum_{i=1}^K \frac{e^{f_i(x)}}{Z(\theta)},
\end{equation}
where $Z(\theta)$ is an unknown partition function. The energy score of a test point $x$ can be then defined as:
\begin{equation} \label{eq:grathwohl2020}
    E(x) = - \log{\sum_{i=1}^K e^{f_i(x)}}.
\end{equation}
Building upon this approach, the recently proposed method by Liu et al. \cite{liu2020a} augments \eqref{eq:grathwohl2020} by temperature scaling:
\begin{equation} \label{eq:liu2020a}
    E(x) = - T \log{\sum_{i=1}^K e^{f_i(x)/T}},
\end{equation}
where $T \in \mathbb{R}_{>0}$ is a temperature scaling parameter. \cite{liu2020a} also proposes to fine-tune pre-trained neural network classifiers to enlarge the energy gap between in and out-of-distribution samples. The neural network's weights are updated by minimizing
\begin{equation}
    \mathbb{E}_{x\sim \mathcal{D}_{\text{in}}^{\text{train}}} \left[- \log \max_j \frac{ e^{{f_j(x)}}}{\sum_{i=1}^K e^{f_i(x)}}, \right] + \lambda \mathcal{L}_\text{energy},
\end{equation}
where
$
    \mathcal{L}_\text{energy} = \mathbb{E}_{x\sim \mathcal{D}_{\text{in}}^{\text{train}}} \left[ \ell_{m_{\text{in}}}(E(x))^2 \right]
    + \mathbb{E}_{x\sim \mathcal{D}_{\text{out}}^{\text{train}}} \left[ \ell_{m_{\text{out}}}(-E(x))^2 \right]$,
$\mathcal{D}_{\text{out}}^{\text{train}}$ is an auxiliary dataset drawn from $Q$, and $\ell_m(x) = \max (x-m, 0)$ is the hinge loss with hyperparameter $m$. Note that, while the fine-tuning improves OOD performance, it worsens the classification accuracy. On common benchmarks, this work is currently considered the state of the art in energy-based OOD detection.
    
Similar to \cite{grathwohl2020} and \cite{liu2020a}, our proposed approach is energy-based. However, we employ a different definition of $E(x)$ that is not based on the output layer of a neural network. Instead, it is the St-RKM's \cite{strkm} objective function that stems from the RKM energy \cite{drkm} and includes a reconstruction loss term. This is described further in the following section.

\section{Proposed Model}
\label{met}
In this section, we discuss the proposed energy-based method for out-of-distribution detection building upon the latent variable model St-RKM \cite{strkm}. 
Consider the objective function of the St-RKM:
\begin{equation} \label{eq:strkm}
J(x) =
\Vert \phi_{\theta}(x) - \bm{U}\bm{h}\Vert_{2}^{2} + \lambda \loss{\xi}{}{}{\theta},%
\end{equation}
with feature map $\phi_\theta(x) \in \mathbb{R}^l$, latent variable $h \in \mathbb{R}^m$ with $m \leq l$, real-valued parameter vectors $\bm{\theta}$  and $\bm{\xi}$, regularization parameter $\lambda>0$ and the interconnection matrix $U = [\bm{u}_1, \dots , \bm{u}_m]$ belonging to the Stiefel manifold $\St(\ell,m)$, that is, the set of $\ell\times m$ matrices with orthonormal columns ($\ell\geq m$). The feature map is  assumed to be centered, i.e., $\mathbb{E}_{\bm{x}\sim p(\bm{x})} [\bm{\phi}_{\bm{\theta}}(\bm{x})] = \bm{0}$.
Note that the St-RKM objective consists of the RKM energy \cite{drkm} with additional regularization terms and a reconstruction loss term $L_{\bm{\xi},U}$ (e.g., the AutoEncoder (AE) loss). 

First, we train the St-RKM on the given training set $\Dtrainin$ in an unsupervised manner. Following \cite{strkm}, this is done by optimizing the sum of the objective \eqref{eq:strkm} over $\Dtrainin$:
\begin{equation} \label{eq:optproblem}
    \min_{\substack{ U\in \St(\ell,m)\\\bm{\theta}, \bm{\xi}}}\min_{\bm{h}_i\in \mathbb{R}^m} \sum_{i=1}^N
    J(x_i),
\end{equation}
where $N = \vert \Dtrainin \vert$.
By minimizing first over $\bm{h}_i$ in \eqref{eq:optproblem}, we find the score vector $\bm{h}_i^\star = U^\top  \bm{\phi}_{\bm{\theta}}(\bm{x}_i)$ with respect to the columns of $U$, the orthonormal set $\{\bm{u}_1, \dots, \bm{u}_m\}$. 
After substitution, the optimization problem becomes
\begin{align}
\min_{\substack{ U\in \St(\ell,m)\\\bm{\theta}, \bm{\xi}}}\frac{1}{N}\sum_{i=1}^{N} \underbrace{ \| (\mathbb{I} - \bm{U}\bm{U}^\top) \phi_{\theta}(x) \|_2^2 }_{ \text{KPCA reconstruction}} + \lambda \underbrace{ L_{\bm{\xi},U}(\bm{x}_i,\bm{\phi}_{\bm{\theta}}(\bm{x}_i)) }_{\text{AutoEncoder loss}},\label{eq:ReducedObjective}
\end{align}
where we use mean-squared error as the autoencoder loss function $L_{\bm{\xi},U}(\bm{x},\bm{\phi}_{\bm{\theta}}(\bm{x})) = \left\|\bm{x} - \bm{\psi}_{\bm{\xi}}\big(\mathbb{P}_U\bm{\phi}_{\bm{\theta}}(\bm{x}) \big)\right\|_2^2$. Here the feature-map $\bm{\phi}_{\bm{\theta}}(\cdot)$ is the encoder map from the input space to the latent space and $\bm{\psi}_{\bm{\xi}}(\cdot)$ represents a decoder map. 

As discussed in \cite{strkm}, a PCA interpretation can be given to the first term of \eqref{eq:ReducedObjective}. By introducing the covariance matrix
$
C_{\bm{\theta}} = \frac{1}{n}\sum_{i=1}^n  \bm{\phi}_{\bm{\theta}}(\bm{x}_i)\bm{\phi}_{\bm{\theta}}^\top(\bm{x}_i)$, the first term in~\eqref{eq:ReducedObjective} can be written as $\Tr\left( C_{\bm{\theta}} - \mathbb{P}_U C_{\bm{\theta}} \mathbb{P}_U\right)
$, which corresponds to the reconstruction error of Kernel PCA for the kernel $k_{\bm{\theta}}(\bm{x},\bm{y}) =  \bm{\phi}^\top_{\bm{\theta}}(\bm{x}) \bm{\phi}_{\bm{\theta}}(\bm{y}) $. If $\mathbb{P}_U= UU^\top$ is the projector on the $m$ principal components, then $U^\top C_{\bm{\theta}} U = \diag(\bm{\lambda})$, where $\bm{\lambda}$ is a vector containing the principal values.

After having trained the St-RKM model, the energy function for out-of-distribution detection is defined as the objective \eqref{eq:strkm} with the learned interconnection matrix $U^\star$, feature map parameter $\bm{\theta^\star}$ and pre-image map parameter $\bm{\xi^\star}$:
\begin{align} \label{eq:energy}
E_{\text{FullEnergy}}(x) =& {\h{}{} - \corr{\theta^\star}{}{\star}{} + \phimap{\theta^\star}{}} 
+ \lambda \loss{\xi^\star}{\star}{}{\theta^\star},
\end{align}
where $\bm{h} = {U^\star}^\top  \bm{\phi}_{\bm{\theta^\star}}(\bm{x})$. In the following discussion, we refer to this energy score as $E_{\text{FullEnergy}}$.

It is expected that for \emph{similar} in/out-distributions, the Autoencoder loss term takes similar values for samples from each case, thus making it harder to distinguish between in- and out-of-distribution samples with the Autoencoder reconstruction error. In such instances, looking at the KPCA  reconstruction term alone might be  more useful. This is illustrated in Fig. \ref{fig:ood_cifar}.
In other words, this suggests that the $E_{\text{FullEnergy}}$ 
score should perform better on OOD datasets whose distribution is further away from the training distribution, while the KPCA reconstruction term could result in better performance on OOD datasets whose distribution is closer to $P$. This motivates investigating the individual components of \eqref{eq:energy} by treating them as standalone energy metrics. %
    Hence, we define the following additional energy function for OOD detection:
\begin{equation} \label{eq:noloss}
E_{\text{kPCAError}}(x) = \h{}{} - \corr{\theta^\star}{}{\star}{} +  \phimap{\theta^\star}{}.
\end{equation}
In other words, the $E_{\text{kPCAError}}$ energy definition is the energy \eqref{eq:energy} without the Autoencoder loss term. Intuitively, \eqref{eq:noloss} can be seen as the norm of the reconstruction error vector between the latent space points and the projected points onto the subspace (see first term in \eqref{eq:ReducedObjective}). %

\begin{table*}[t]
\caption{Comparison of OOD detection performance. Lower scores ({\small $\downarrow$}) are better for FPR95 and higher scores ({\small $\uparrow$}) are better for AUROC and AUPR. [S] Supervised / [U] Unsupervised. Experiments are repeated 10 times. PCA detection has no randomness involved.\vspace{2.1mm}}

    \centering
    \begin{tabular}{ll llllll}
    \multicolumn{8}{c}{$\Dtrainin$: Fashion-MNIST [Mean (Std) over 10 iterations, values in \%]} \vspace{1mm} \\ \toprule
   \multicolumn{1}{c}{\multirow{2}{*}{$\Dtestout$}} & \multirow{2}{*}{Metric} & \multicolumn{2}{c}{$\St$-RKM variants [U]} & \multirow{2}{*}{Liu2020 [S]} & \multicolumn{1}{c}{\multirow{2}{*}{PCA [U]}} & \multicolumn{1}{c}{\multirow{2}{*}{VAE [U]}} & \multicolumn{1}{c}{\multirow{2}{*}{GAN [U]}}\\ \cmidrule{3-4}
     & & \multicolumn{1}{c}{$E_{\text{FullEnergy}}$} & \multicolumn{1}{c}{$E_{\text{kPCAError}}$} & && \\ \midrule
       \multirow{3}{*}{MNIST} & FPR95\small{($\downarrow$)}& 75.73  (2.7) & \textbf{0.38}  (0.2)  & 75.97  (11.1) & 99.99   & 2.67  (1.0)  & 97.31  (4.5)  \\
                              & AUROC\small{($\uparrow$)}&   69.44  (1.5) & \textbf{99.70}  (0.1) & 78.05  (4.9)  & 73.17   & 99.36  (0.1) & 49.32  (17.7) \\
                                & AUPR\small{($\uparrow$)}  &66.26  (3.0) & \textbf{99.75}  (0.1) & 80.36  (4.0)  & 83.73   & 99.44  (0.1) & 60.75  (15.5)\\ \midrule
    \multirow{3}{*}{dSprites}   & FPR95\small{($\downarrow$)}& 99.21  (0.8) & \textbf{2.71}  (2.7)  & 96.23  (3.6) & 99.79   & 69.43  (2.3) & 73.54  (34.8) \\
                                 & AUROC\small{($\uparrow$)}&  11.61  (3.1) & \textbf{99.17}  (0.4) & 63.68  (7.5) & 82.81   & 85.77  (0.8) & 58.04  (31.9) \\
                                   & AUPR\small{($\uparrow$)}& 0.71  (0.02)  & \textbf{92.82}  (2.9) & 20.82  (8.2) & 70.89   & 36.87  (6.1) & 22.76  (29.4)\\ \midrule
    \multirow{3}{*}{SVHN}  & FPR95\small{($\downarrow$)} & \textbf{1.34}  (0.2) & 28.64  (10.1) & 29.14  (8.9) & 75.31  & 27.42  (6.1) & 82.55  (7.2)  \\
                             & AUROC\small{($\uparrow$)} & \textbf{99.59} (0.04) & 95.61  (1.3) & 94.04  (2.0) & 51.36  & 94.56  (1.3) & 59.99  (5.0) \\
                               & AUPR\small{($\uparrow$)}& \textbf{99.23} (0.1) & 93.00  (1.8) & 88.52  (3.3) & 25.57  & 89.76  (2.4) & 42.15  (10.7)  \\ \midrule
    \multirow{3}{*}{CIFAR-10}   & FPR95\small{($\downarrow$)} & \textbf{0.34}  (0.01)  & 13.40  (5.7) & 46.97  (10.4) & 65.76   & 6.50  (2.8)  & 81.09  (7.9) \\
                                  & AUROC\small{($\uparrow$)}& \textbf{99.76}  (0.003) & 97.70  (0.8) & 90.60  (2.5)  & 67.86   & 98.63  (0.4) & 69.32  (6.4) \\
                                  & AUPR\small{($\uparrow$)} & \textbf{99.83}  (0.003) & 98.08  (0.6) & 91.77  (2.0)  & 60.69   & 98.83  (0.3) & 73.30  (3.9)
\\\bottomrule
    \end{tabular}
	\label{tab:res1}
\end{table*}

\section{Experimental Evaluation} \label{exp}
In this section, we discuss the experiments on standard computer vision datasets and on a time series problem for anomalous heartbeat detection. 

\textbf{Training details.} We train all models for 1600 epochs with mini-batch size of 256. For Liu et al. \cite{liu2020a}, VAE and GAN methods, training is performed using Adam \cite{KingmaAdam} with a learning rate of $2 \times 10^{-4}$. For St-RKM, an alternating minimization training scheme is used: the weights of the encoder $\bm{\phi}_{\bm{\theta}}(\cdot)$ and of the decoder $\bm{\psi}_{\bm{\xi}}(\cdot)$ are jointly optimized with the Adam optimizer \cite{KingmaAdam} with learning rate $2 \times 10^{-4}$, and the interconnection matrix $U$ is optimized with the Cayley Adam optimizer \cite{Li2020Efficient} with learning rate $1 \times 10^{-4}$. For the full training algorithm, see Algorithm 1 in \cite{strkm}. A PCA baseline is also considered; in this case, scores are computed by the reconstruction error by removing the components whose contribution to total variation is less than 2\%. The subspace dimension is set to 10 for all datasets. The source code and additional architectural details are available at \url{https://github.com/taralloc/st-rkm-ood}. %

\textbf{Performance metrics.} We evaluate the false positive rate (FPR95) at 95\% true positive rate, area under the receiver operating characteristic curve (AUROC) and under the precision-recall curve (AUPR). Note that the threshold $\gamma$ is not needed for evaluating AUROC and AUPR scores. We also evaluate the Overlapping Coefficient to quantify the dissimilarity between the $\Dtrainin$ and $\Dtestout$ energy distributions. For two real probability density functions $f_A (x)$ and $f_B (x)$, the overlapping coefficient \cite{inman1989,pastore2019} $\eta \colon \mathbb{R}^n \times \mathbb{R}^n \rightarrow [0, 1]$ is defined as:
\begin{equation}
    \eta(A, B) = \int_{\mathbb{R}^n} \min [f_{A}(x), f_{B}(x) ] dx .
    \label{eq:overlap_coefficient}
\end{equation}
$\eta (A, B) = 0$ indicates that the support of $f_A (x)$ and $f_B (x)$ does not have any common points, i.e., the distributions do not overlap. Ideally, this should be the case for $\Dtrainin$ and $\Dtestout$ energy distributions. 

\subsection{Computer Vision}
First, we describe our experimental setup in Section \ref{exp:cv:setup}. In Section \ref{exp:cv:exp}, we show that the proposed method improves over the state of the art in energy-based OOD detection using a pre-trained neural classifier on the considered benchmarks. We also show the distribution of the energy scores on multiple OOD datasets.

\subsubsection{Experimental Setup} \label{exp:cv:setup}
For this set of experiments, we create a Convolutional Network architecture for the St-RKM's encoder $\bm{\phi}_{\bm{\theta}}(\cdot)$ and a transposed Convolutional architecture for the St-RKM's decoder $\bm{\psi}_{\bm{\xi}}(\cdot)$.
\begin{itemize}
    \item \emph{In-distribution  and  out-of-distribution  datasets}. We use Fashion-MNIST from \cite{fashionmnist} as $\mathcal{D}^{\text{train}}_{\text{in}}$. For testing, we use four different datasets: MNIST \cite{mnist}, dSprites \cite{betavae}, SVHN \cite{svhn} and CIFAR-10 \cite{cifar10}. SVHN, dSprites and CIFAR-10 are resized to $28\times 28 \times 1 $ input dimensions.
    
    \item \emph{Hyperparameters.} Following a sensitivity analysis, in \eqref{eq:energy} we fix $\lambda = 100$. The St-RKM, VAE and GAN models employ the same encoder-decoder architecture; for GAN, the encoder is used in the discriminator, while the generator uses the decoder. The encoder $\bm{\phi}_{\bm{\theta}}$ consists of three convolutional (\textit{Conv}) layers with doubling channel size followed by two fully connected (\textit{FC}) layers, and the decoder $\bm{\psi}_{\bm{\xi}}$ consists of two \textit{FC} layers followed by three transposed convolutional (\textit{ConvTr}) layers with halving channel size. All convolutions have stride 2 and padding 1, except for the last \textit{Conv} layer of $\bm{\phi}_{\bm{\theta}}$ and for the first \textit{ConvTr} layer of $\bm{\psi}_{\bm{\xi}}$ that have stride 1 and no padding. The activation function is Parametric-RELU ($\alpha = 0.2$), except for the output layer of $\bm{\psi}_{\bm{\xi}}$ that has Sigmoid activation function. For \cite{liu2020a}, we train a convolutional neural network classifier whose architecture is set to be identical to the architecture of the St-RKM's encoder.
\end{itemize}

\subsubsection{Experimental Results} \label{exp:cv:exp}

\begin{figure*}[t!]
	\centering
	\includegraphics[width=.8\textwidth]{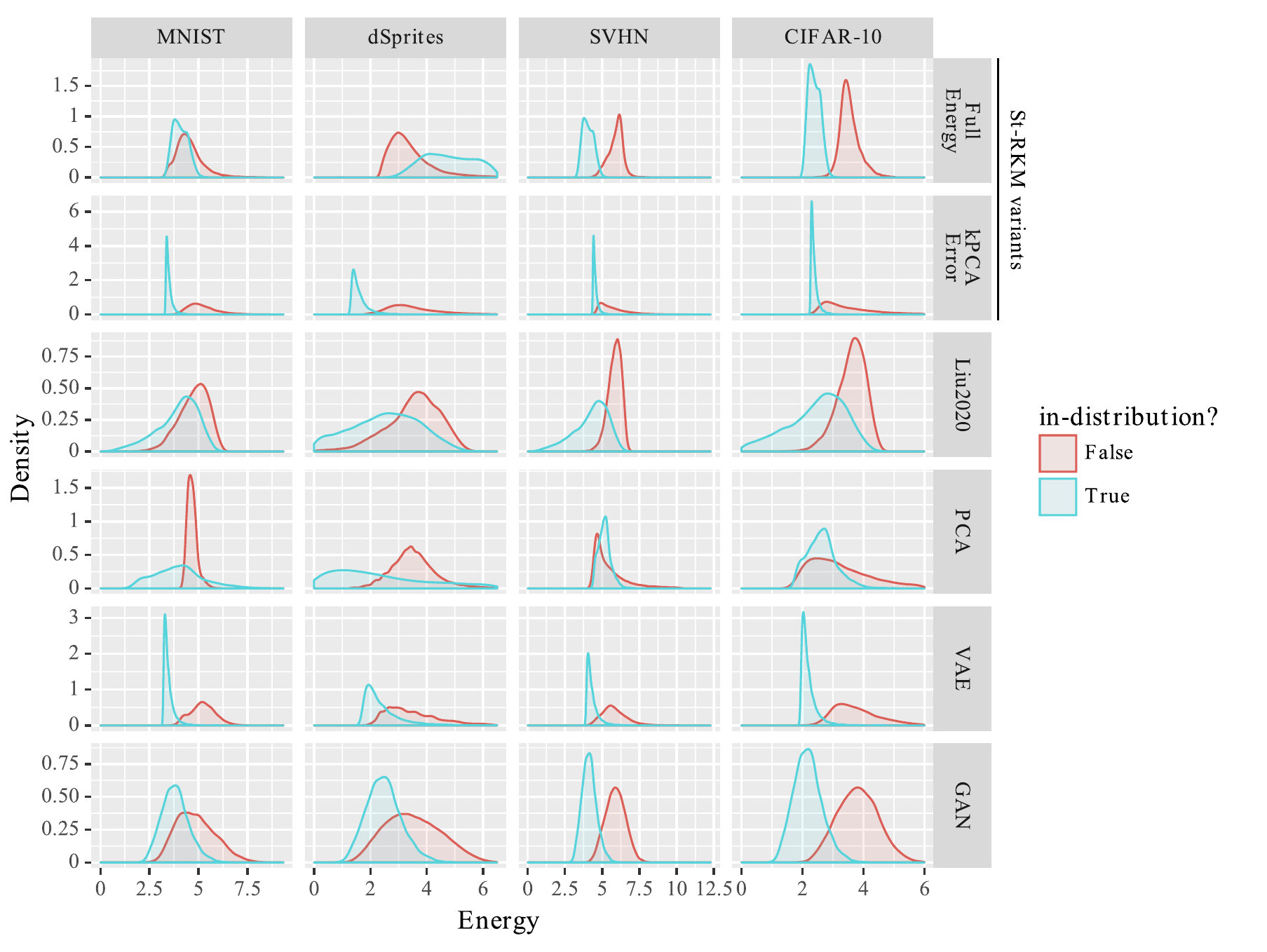}
	\caption{Visualizing the distribution of the energy scores of various models. In energy-based models, low energies are associated with more likely input samples, hence it corresponds to in-distribution samples. In line with the properties of each energy definition discussed in Section \ref{met}, note that, for MNIST and dSprites, the in- and out-distributions overlap for the $E_{\text{FullEnergy}}$ definition, but they are separated for the $E_{\text{kPCAError}}$ definition, suggesting that in those datasets the autoencoder loss term has a dominant weight in the energy function of the $E_{\text{FullEnergy}}$ definition. In general, the distributions of the best-performing methods based on the St-RKM's energy look smoother and more separated than the distribution of the method proposed in \cite{liu2020a}.}
	\label{fig:plot1}
\end{figure*}

\begin{table*}[]
\centering
\caption{Overlapping Coefficient quantifying the overlap of $\Dtrainin$ and $\Dtestout$ energy distribution as shown in Fig. \ref{fig:plot1}. The standard deviation over 10 iterations is given in parenthesis. Smaller is better.}
\label{tab:overlap_coefficient}
\begin{tabular}{lllllll}
\toprule
    \multirow{2}{*}{$\Dtestout$} & \multicolumn{2}{c}{$\St$-RKM variants} & \multirow{2}{*}{Liu2020} & \multirow{2}{*}{PCA} & \multirow{2}{*}{VAE} & \multirow{2}{*}{GAN}\\ \cmidrule{2-3}
     &  $E_{\text{FullEnergy}}$ & $E_{\text{kPCAError}}$ & && \\ \midrule
MNIST    & 0.58  (0.02) & \textbf{0.03}  (0.003) & 0.41  (0.1) & 0.24  & 0.04  (0.005) & 0.43  (0.1)     \\
dSprites & 0.24  (0.05) & \textbf{0.04}  (0.01) & 0.49  (0.1) & 0.19  & 0.28  (0.01) & 0.31  (0.2) \\
SVHN     &  \textbf{0.03} (0.003) & 0.12  (0.02) & 0.17  (0.04) & 0.52  & 0.14  (0.02) & 0.60  (0.1) \\
  CIFAR-10 & \textbf{0.01} (0.001) & 0.09  (0.01) & 0.22  (0.04) & 0.51  & 0.06  (0.01) & 0.51  (0.1) \\
\bottomrule
\end{tabular}
\end{table*}

\begin{figure}[t!]
	\centering
	\includegraphics[width=0.5\textwidth]{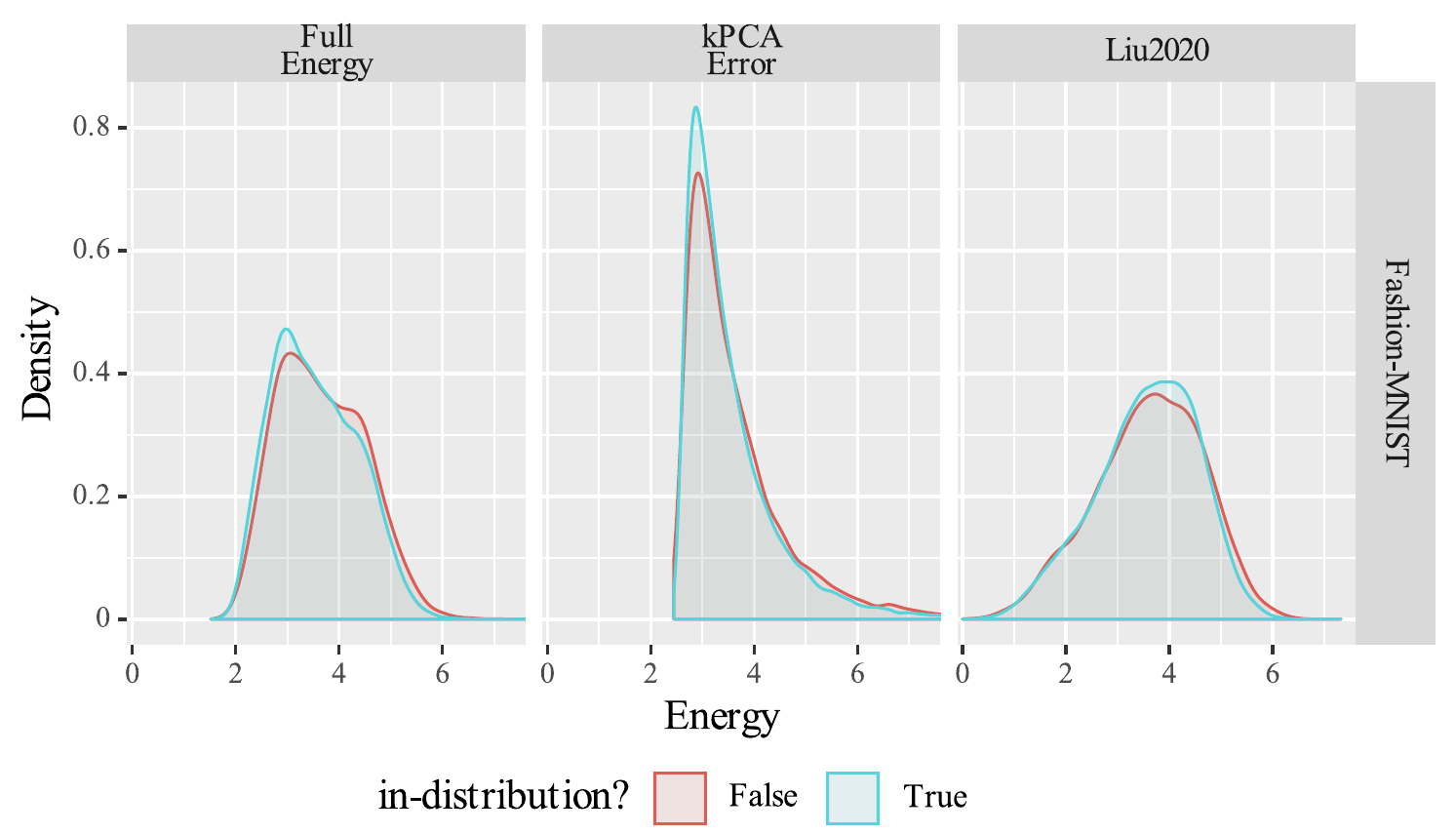}
	\caption{Distribution of the energy scores when the in/out-distributions are from the same dataset. Here, the $\Dtrainin$ is the training set of Fashion-MNIST ($N_{\text{train}} = 60000$) and $\Dtestout$ is the test set ($N_{\text{test}} = 10000$) of the same dataset.}
	\label{fig:sanitycheck}
\end{figure}

\begin{figure}[t!]
	\centering
    \setlength{\tabcolsep}{0.3pt}
	\begin{tabular}{cccccccccc}
		\begin{minipage}{.047\textwidth}
			\includegraphics[width=\textwidth]{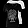}
		\end{minipage}{} &
		\begin{minipage}{.047\textwidth}
			\includegraphics[width=\textwidth]{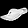}
		\end{minipage}{} &
		\begin{minipage}{.047\textwidth}
			\includegraphics[width=\textwidth]{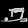}
		\end{minipage}{} &
		\begin{minipage}{.047\textwidth}
			\includegraphics[width=\textwidth]{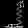}
		\end{minipage}{} &
		\begin{minipage}{.047\textwidth}
			\includegraphics[width=\textwidth]{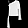}
		\end{minipage}{} &
		\begin{minipage}{.047\textwidth}
			\includegraphics[width=\textwidth]{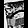}
		\end{minipage}{} &
		\begin{minipage}{.047\textwidth}
			\includegraphics[width=\textwidth]{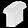}
		\end{minipage}{} &
		\begin{minipage}{.047\textwidth}
			\includegraphics[width=\textwidth]{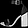}
		\end{minipage}{} &
		\begin{minipage}{.047\textwidth}
			\includegraphics[width=\textwidth]{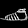}
		\end{minipage}{} &
		\begin{minipage}{.047\textwidth}
			\includegraphics[width=\textwidth]{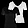}
		\end{minipage}{} \\[3mm]
	
		\begin{minipage}{.047\textwidth}
			\includegraphics[width=\textwidth]{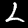}
		\end{minipage}{} &
		\begin{minipage}{.047\textwidth}
			\includegraphics[width=\textwidth]{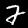}
		\end{minipage}{} &
		\begin{minipage}{.047\textwidth}
			\includegraphics[width=\textwidth]{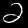}
		\end{minipage}{} &
		\begin{minipage}{.047\textwidth}
			\includegraphics[width=\textwidth]{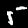}
		\end{minipage}{} &
		\begin{minipage}{.047\textwidth}
			\includegraphics[width=\textwidth]{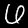}
		\end{minipage}{} &
		\begin{minipage}{.047\textwidth}
			\includegraphics[width=\textwidth]{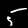}
		\end{minipage}{} &
		\begin{minipage}{.047\textwidth}
			\includegraphics[width=\textwidth]{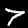}
		\end{minipage}{} &
		\begin{minipage}{.047\textwidth}
			\includegraphics[width=\textwidth]{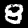}
		\end{minipage}{} &
		\begin{minipage}{.047\textwidth}
			\includegraphics[width=\textwidth]{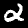}
		\end{minipage}{} &
		\begin{minipage}{.047\textwidth}
			\includegraphics[width=\textwidth]{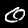}
		\end{minipage}{} \\[3mm]
		
		\begin{minipage}{.047\textwidth}
			\includegraphics[width=\textwidth]{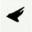}
		\end{minipage}{} &
		\begin{minipage}{.047\textwidth}
			\includegraphics[width=\textwidth]{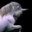}
		\end{minipage}{} &
		\begin{minipage}{.047\textwidth}
			\includegraphics[width=\textwidth]{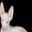}
		\end{minipage}{} &
		\begin{minipage}{.047\textwidth}
			\includegraphics[width=\textwidth]{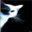}
		\end{minipage}{} &
		\begin{minipage}{.047\textwidth}
			\includegraphics[width=\textwidth]{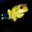}
		\end{minipage}{} &
		\begin{minipage}{.047\textwidth}
			\includegraphics[width=\textwidth]{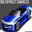}
		\end{minipage}{} &
		\begin{minipage}{.047\textwidth}
			\includegraphics[width=\textwidth]{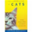}
		\end{minipage}{} &
		\begin{minipage}{.047\textwidth}
			\includegraphics[width=\textwidth]{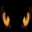}
		\end{minipage}{} &
		\begin{minipage}{.047\textwidth}
			\includegraphics[width=\textwidth]{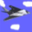}
		\end{minipage}{} &
		\begin{minipage}{.047\textwidth}
			\includegraphics[width=\textwidth]{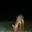}
		\end{minipage}{} \\

	\end{tabular}
	\caption{Samples from the test set of Fashion-MNIST (first row), MNIST (second row), and CIFAR-10 (third row) that are flagged as OOD by our method, illustrating that these samples often show unusual features.}
	\label{fig:outliers}
\end{figure}

\textbf{Can the model detect samples from the same distribution?} First, we investigate the extreme case where the in-distribution $P$ is the same as the out-distribution $Q$. We consider the training and test sets of Fashion-MNIST: the training set is used as $\Dtrainin$ to train the OOD detectors, and the test set is taken as $\Dtestout$ to evaluate OOD performance. A significant overlap, as can be seen in  Fig. \ref{fig:sanitycheck}, suggests that the model performs as expected and overall captures the true positives. Some samples that are flagged as OOD are shown in Fig. \ref{fig:outliers}. Interestingly, most of such instances are even difficult for a human to classify. 

\textbf{How does the proposed method perform compared to its competitors?}  Table \ref{tab:res1} shows the performance attained by various energy function definitions and \cite{liu2020a}'s energy function without fine-tuning, evaluated for each out-of-distribution dataset. In these benchmarks, our proposed method outperforms the current state-of-the-art energy-based out-of-distribution detection method on all considered OOD datasets. For instance, on MNIST our proposed $E_{\text{kPCAError}}$ score achieves 0.38\% FPR compared to 75.97\% FPR achieved by \cite{liu2020a}. On CIFAR-10, our proposed $E_{\text{FullEnergy}}$ energy definition achieves 0.34\% FPR compared to 46.97\% FPR achieved by \cite{liu2020a}.

\textbf{Which component of the Energy function is more useful?} Table \ref{tab:res1} further gives an insight regarding the merit of various components of the energy function as discussed in Section \ref{met}.  It shows that, on the one hand, the $E_{\text{FullEnergy}}$ score performs better on OOD datasets whose distribution is further away from the training distribution $P$; this is the case for SVHN and CIFAR-10. On the other hand, the $E_{\text{kPCAError}}$ score performs best on OOD datasets whose distribution is closer to $P$; this is the case for MNIST and dSprites. 
Note that the $E_{\text{FullEnergy}}$ performs poorly in the latter case also, as can be seen from Fig. \ref{fig:plot1}. Fig. \ref{fig:plot1} shows a kernel density plot with Gaussian kernel over the energy distributions for each method (row) and $\Dtestout$ (column). First, the energy scores are mean centered, scaled to unit variance, and finally shifted by adding the negative minimum energy score. The subplots of the $E_{\text{FullEnergy}}$ definition for MNIST and dSprites show significant overlap, and thus the energy gap between in-distribution samples and OOD samples is small. It can be understood as follows. Note that the AE loss of \eqref{eq:energy} dominates over the kernel PCA reconstruction in $E_{\text{FullEnergy}}$ due to the significantly large number of neural network parameters compared to that in $U$. Further, the AE loss of samples from $P$ is similar to that of samples from $Q$ since the network's parameters would be similar if trained on similar distributions. In such instances, looking at the $E_{\text{FullEnergy}}$ naively obfuscates the distinction between in/out-of-distribution samples. However, $E_{\text{kPCAError}}$ highlights the projection error between similar datasets and therefore results in better OOD detection performance.

\textbf{How well does the Energy function separates the in/out distributions?} Table \ref{tab:overlap_coefficient} shows the overlapping coefficient for the considered datasets and OOD detectors, further confirming that the proposed energy functions based on the St-RKM's energy can lead to better performance than both generative-based and discriminative-based methods. The St-RKM variants give the lowest overlap, meaning that they induce energy distributions effectively separating $\Dtrainin$ samples from $\Dtestout$ samples. This performance increase is especially noteworthy given that our method does not use any label information in any way, contrary to \cite{liu2020a}, whose energy function \eqref{eq:liu2020a} is defined based on a classifier trained in the supervised learning setting. At the same time, the fine-tuning procedure with auxiliary OOD training data proposed in \cite{liu2020a} could boost the performance of their method; however, in many applications, access to OOD samples during training is limited. In contrast, the training procedure of our method does not require additional fine-tuning with OOD training data.

\subsection{Time Series Data}
We now turn to OOD detection in time series data from the healthcare sector. This is an increasingly important  area where unsupervised OOD detection algorithms 
can be extremely useful, as obtaining and labeling medical data is usually expensive. The proposed method can be applied to time series data by selecting an appropriate encoder-decoder architecture for the St-RKM. On the contrary, the method proposed in \cite{liu2020a} is not directly applicable to time series data because its score function is defined on the output layer of a neural network classifier.

\subsubsection{Experimental Setup} \label{exp:time:setup}
For this set of experiments, the encoder-decoder architecture of St-RKM, VAE, and GAN models is parametrized by a Variational Recurrent AutoEncoder (VRAE) \cite{fabius2015}.
\begin{itemize}
    \item \emph{In-distribution and out-of-distribution datasets.} We use the publicly available ECG5000 dataset \cite{ecg5000}, which consists of $N=5000$ electrocardiogram (ECG) univariate time series. Each sequence has 140 timesteps and corresponds to a single heartbeat. The dataset contains 5 classes: one class is seen as \textit{normal}, while the others represent anomalous heartbeats. The dataset is first divided into a training set of $N_{\text{train}}=4500$ sequences and a test set of $N_{\text{test}}=500$ sequences. The sequences of the \textit{normal} class contained in the training set are taken as $\Dtrainin$, and the sequences of the anomalous heartbeats in the test set are taken as $\Dtestout$.
    
    \item \emph{Hyperparameters.} In all models, following the architecture proposed in \cite{fabius2015}, the encoder uses an LSTM with two layers, and the decoder uses an LSTM with two layers followed by a fully connected layer. All LSTM layers have input size 1 and hidden size 10.
    
\end{itemize}

\subsubsection{Experimental Results}

\begin{table}[t]
\renewcommand{\arraystretch}{1.06}
\caption{Comparison of OOD detection performance in time series data of electrocardiogram (ECG) sequences. All values are in percentages. \label{tab:res2}}
\centering
\begin{tabular}{l cccc}
\toprule
\multirow{2}{*}{Metric} & \multicolumn{2}{c}{$\St$-RKM variants} & \multirow{2}{*}{VRAE} & \multirow{2}{*}{GAN}\\ \cmidrule{2-3}
     &  $E_{\text{FullEnergy}}$ & $E_{\text{kPCAError}}$ & & \\ \midrule
FPR95\small{($\downarrow$)}&  \textbf{6.18} (0.2)  & 98.45 (1.7) & \textbf{6.27}  (0.3)  & 87.45 (19.1) \\
AUROC\small{($\uparrow$)}&  \textbf{94.02}  (0.1) & 50.39  (1.8)& 93.89  (0.2) & 37.08  (29.9)\\
AUPR\small{($\uparrow$)} &  \textbf{95.62}  (0.2) & 85.67  (0.2)& \textbf{95.71}  (0.2) & 78.94  (9.6)\\
\bottomrule
\end{tabular}
\end{table}

\textbf{Is the proposed method able to detect anomalous heartbeats?}  Table \ref{tab:res2} shows the performance attained by various proposed energy function definitions, by VAE with VRAE architecture, and by GAN. When computing the scores, all anomalous classes are seen as one out-of-distribution class.  The $E_{\text{FullEnergy}}$ score is the highest-performing energy function among the St-RKM variants. It  outperforms GAN models on all metrics, providing more reliable performance as well.
A possible explanation for the poor performance of GAN models is that the discriminator is trained to differentiate between real heartbeat sequences and sequences generated by the generator, rather than to distinguish between real and anomalous heartbeats. Hence, GANs model only $P$ implicitly. Compared to VAE, the $E_{\text{FullEnergy}}$ score has similar performance with slightly better average FPR and AUROC. Overall, our proposed method can distinguish anomalous heartbeats with an average FPR of 6.18\% and an average AUROC of 94.02\%.

\section{Conclusion}
In this work, we proposed an energy-based method for the out-of-distribution detection task based on the St-RKM model. Further, we proposed multiple energy functions, examined their properties, and discussed practical scenarios where each proposed definition may perform best. In contrast to the state of the art in energy-based OOD detection which exploits a pre-trained classifier, our method does not require labels as it is trained in an unsupervised manner. Moreover, the training phase of our method does not require anomalous data, contrary to previous methods that use a fine-tuning step with an auxiliary dataset of OOD samples. We evaluated our method on several out-of-distribution datasets and showed that it improves on the baseline (state of the art in energy-based OOD detection) of \cite{liu2020a}. We showed that the proposed $E_{\text{kPCAError}}$ definition is more suitable when the out-distribution is similar to the training distribution, while the proposed $E_{\text{FullEnergy}}$ definition performs best when the out-distribution is further away from the training distribution. Future work includes investigating the proposed method against adversarial attacks.

\section*{Acknowledgment}
{\footnotesize
EU: The research leading to these results has received funding from the European Research Council under the European Union’s Horizon 2020 research and innovation program/ERC Advanced Grant E-DUALITY (787960). This paper reflects only the authors’ views and the Union is not liable for any use that may be made of the contained information. Research Council KUL: Optimization frameworks for deep kernel machines C14/18/068. Flemish Government: FWO: projects: GOA4917N (Deep Restricted Kernel Machines: Methods and Foundations), Ph.D./Postdoc grant Impulsfonds AI: VR 2019 2203 DOC.0318/1QUATER Kenniscentrum Data en Maatschappij. Ford KU Leuven Research Alliance Project KUL0076 (Stability analysis and performance improvement of deep reinforcement learning algorithms). EU H2020 ICT-48 Network TAILOR (Foundations of Trustworthy AI - Integrating Reasoning, Learning and Optimization). This work was supported by the Research Foundation Flanders (FWO) research projects G086518N, G086318N, and G0A0920N; Fonds de la Recherche Scientifique — FNRS and the Fonds Wetenschappelijk Onderzoek — Vlaanderen under EOS project no 30468160 (SeLMA). Johan Suykens and Panagiotis Patrinos are affiliated to Leuven.AI - KU Leuven institute for AI, B-3000, Leuven, Belgium. The computational infrastructure and services used in this work were provided by the VSC (Flemish Supercomputer Center), funded by the Research Foundation - Flanders (FWO) and the Flemish Government.
}

\bibliographystyle{IEEEtran}
\bibliography{references}

\clearpage
\appendix

\section{Architecture and Training Details} \label{app:arch}
We train all models for 1600 epochs with mini-batch size of 256. For \cite{liu2020a}, VAE and GAN methods, training is performed using Adam \cite{KingmaAdam} with a learning rate of 0.0002. For St-RKM, an alternating minimization scheme using the Adam \cite{KingmaAdam} and Cayley Adam \cite{Li2020Efficient} optimizer is employed; the learning rate of the optimizers is set to 0.0002 and 0.0001, respectively. Moreover, for the GAN method, in \eqref{eq:gan} we fix $\lambda = 0$, as this choice led to the best AUROC performance in the experimental evaluation of \cite{skvara2018}. The PCA baseline scores are computed by the reconstruction error when components whose contribution to total variation is less than 2\% are removed. The subspace dimension is set to 10 for all employed datasets.

The model architectures for the Computer Vision experiments are shown in Table \ref{tab:archcv}. All convolutions (\textit{Conv}) and transposed convolutions (\textit{ConvTr}) are with stride 2 and padding 1, unless stated otherwise. Layers have Parametric-RELU ($\alpha = 0.2$) activation functions, except the output layer of the pre-image map $\psi_{\zeta}$ that has Sigmoid activation function (since input data is normalized $[0,1]$). The last convolutional layer of $\phi_{\theta}$ and the first transposed convolutional layer of $\psi_{\xi}$ have stride 1 and no padding. SVHN, dSprites and CIFAR-10 are resized to $28\times 28 \times 1 $ input dimensions.

\begin{table}[h]
\caption{Model architectures for the Computer Vision experiments. For all, $c=40$ and $\hat{k}=3$.  \label{tab:archcv}}
\centering
	\begin{tabular}{@{} l l @{}}
		\toprule
		\multicolumn{2}{c}{\textbf{Architecture} ($\Dtrainin$: Fashion-MNIST)}                        \\
		\midrule
		\hspace{-1mm}
	    $\bm \phi_{\theta}(\cdot) = \begin{cases} \makecell[l]{Conv~ [c]\times 4 \times 4; \\Conv ~[c \times 2] \times 4 \times 4;\\ Conv~ [c \times 4] \times \hat{k}\times\hat{k};     \\  FC~ 256; \\ FC~ 50 ~(Linear)}    \end{cases} $ & \hspace{-6.5mm} $ \bm \psi_{\zeta}(\cdot) = \begin{cases}
		FC~256;\\
		FC~ [c \times 4] \times \hat{k}\times\hat{k};\\
		ConvTr ~[c \times 4] \times 4 \times 4;\\
		ConvTr ~[c \times 2] \times 4 \times 4;\\
		ConvTr ~[c]~ (Sigmoid)\\
		\end{cases} $ \hspace{-3mm} \\
		\bottomrule
	\end{tabular}
\end{table}

\begin{table}[h]
\caption{Model architectures for the Time Series experiments. This encoder-decoder architecture is called Variational Recurrent AutoEncoder \cite{fabius2015}. All LSTM layers have input size 1 and hidden size 10.}
\centering
\begin{tabular}{l l}
\toprule
\multicolumn{2}{c}{\textbf{Architecture} ($\Dtrainin$: ECG5000)} \\ \midrule
$\bm \phi_{\theta}(\cdot) = \begin{cases} LSTM~\text{(2 layers)} \end{cases}$ & $\bm  \psi_{\zeta}(\cdot) = \begin{cases} LSTM~\text{(2 layers)}\\ FC~1 \end{cases}$   \\ \bottomrule       
\end{tabular}
\end{table}

\begin{table}[h]
	\caption{Datasets and hyperparameters used for the experiments. $N$  is the number of instances, $d$ the input dimension, $m$ the subspace dimension and $M$ the minibatch size.}
	\label{Table:dataset}
	\centering
	\begin{tabular}{lccccc}
		\toprule
		\textbf{Dataset}       & $N$    & $d$                       & $m$       & $M$ \\ \midrule
		MNIST                  & 60000  & $28 \times 28$            & 10        & 256 &\\
		Fashion-MNIST          & 60000  & $28 \times 28$            & 10        & 256 &\\
		SVHN                   & 73257  & $32 \times 32 \times 3$   & 10        & 256 &\\
		dSprites               & 737280 & $64 \times 64$            & 10        & 256 &\\
		CIFAR-10               & 60000  & $32 \times 32 \times 3$   & 10        & 256 & \\
		ECG5000                & 5000   & $140$                     & 10        & 256 & \\
		\bottomrule
	\end{tabular}
\end{table}

\FloatBarrier
\section{Additional Energy Definitions} \label{app:energies}
From the various terms $E_{\text{FullEnergy}}$ energy definition of \eqref{eq:energy}, multiple interesting energy definitions can be derived. 
In particular, we also define an energy function consisting of the reconstruction error only, which is a typical approach in OOD detection using deep generative models:
\begin{equation} \label{eq:loss}
E_{\text{AELoss}}(x) = \loss{\xi^\star}{\star}{}{\theta^\star}.
\end{equation}
For large enough $\lambda$, this energy definition is expected to perform similarly to the $E_{\text{FullEnergy}}$. The \textit{AELoss} definition is anticipated to perform best when the distance between the out-distribution and the in-distribution is large, because in this case the reconstruction error of samples from $Q$ is likely to be high.\\

Finally, we define an energy function consisting of the negative correlation term:
\begin{equation} \label{eq:negcorr}
E_{\text{NegCorr}}(x) = \corr{\theta^\star}{}{\star}{}.
\end{equation}
We call this energy definition the \textit{NegCorr} energy. Overall, \eqref{eq:energy} can be seen as:
\begin{align}
E_{\text{FullEnergy}}(x) = \overbrace{\h{}{} - \underbrace{\corr{\theta^\star}{}{\star}{}}_{E_{\text{NegCorr}}} +  \phimap{\theta^\star}{}}^{E_{\text{kPCAError}}}+ \lambda \underbrace{\loss{\xi^\star}{\star}{}{\theta^\star}}_{E_{\text{AELoss}}}.
\end{align}
We report the experimental evaluation of these additional energy definitions in Appendix \ref{app:exp}.

\section{Additional Empirical Results} \label{app:exp}

Complementary to the Table \ref{tab:overlap_coefficient}, Table \ref{tab:mmdwd} shows the 1-Wasserstein distance (WD) and the Maximum Mean Discrepancy (MMD) with Gaussian RBF kernel $k(x,y) = \exp \left( -\frac{\|x-y\|^2_2}{2\sigma^2} \right)$ of the $\Dtrainin$ and $\Dtestout$ energy distributions. For MMD, we set $2\sigma^2$ to the mean of all Euclidean distances between all the scores. As expected, both distance measures tend to be higher for better performing OOD detection methods. For instance, on CIFAR-10 MMD and WD are highest for the $E_{\text{FullEnergy}}$ definition, which in Table \ref{tab:res1} was shown to give the lowest FPR95. 
\begin{table*}[]
\centering
\caption{Maximum Mean Discrepancy (MMD) and 1-Wasserstien distance (WD) between $\Dtrainin$ and $\Dtestout$ energy distributions as determined by various models. Larger is better since it indicates that the distributions are dissimilar and, hence, better suited to OOD detection.}
\label{tab:mmdwd}
\begin{tabular}{llllllll}
\toprule
    \multirow{2}{*}{$\Dtestout$} & Diverg. & \multicolumn{2}{c}{$\St$-RKM variants} & \multirow{2}{*}{Liu2020} & \multirow{2}{*}{PCA} & \multirow{2}{*}{VAE} & \multirow{2}{*}{GAN}\\ \cmidrule{3-4}
     &  & $E_{\text{FullEnergy}}$ & $E_{\text{kPCAError}}$ & && \\ \midrule
\multirow{2}{*}{MNIST} & MMD & 0.071 (0.011) & 0.916 (0.037) & 0.221    (0.071) & 0.31      &  \textbf{1.032} (0.02)  & 0.15  (0.1)\\
                       & WD  & 0.39	(0.039) & 1.657 	(0.03)&	0.979 	(0.15)  &	1.019 	&	\textbf{1.745}	(0.012)&	0.678 (0.229) \\   \midrule
\multirow{2}{*}{dSprites} & MMD & 0.461 (0.096) & \textbf{0.822} (0.093) & 0.133 (0.058) & 0.613  & 0.312 (0.022) & 0.369 (0.265)\\
                           & WD & \textbf{1.886} (0.523) & \textbf{1.871} (0.18)  & 0.989 (0.319) & 3.751  &  1.124 (0.051) & 1.989 (1.671)  \\\midrule
\multirow{2}{*}{SVHN} & MMD  &  \textbf{1.11} (0.023) &  0.5 (0.066) &    0.622  (0.082) & 0.042    & 0.571  (0.066) & 0.046   (0.032) \\
                      & WD &  \textbf{1.845} (0.014)  & 1.289 (0.085) & 1.627 (0.074)  & 0.468   & 1.385 (0.073)  & 0.481 (0.171)\\
\midrule
\multirow{2}{*}{CIFAR-10} & MMD& \textbf{0.903} (0.033) &  0.552 (0.072) & 0.483 (0.064) & 0.102   & 0.733 (0.054) & 0.107  (0.038) \\
                         & WD & \textbf{1.225} (0.029) & \textbf{1.327} (0.104) & \textbf{1.361} (0.072) & 0.716    & 1.539 (0.041) & 0.73  (0.178)\\  \bottomrule
\end{tabular}
\end{table*}

\begin{table*}
	\caption{OOD detection performance for the same-distribution detection experiment. All values are in percentages.}
    \label{tab:sanitycheck}
\centering
\begin{tabular}{lcccccc}
\toprule
\multirow{2}{*}{Metric} &\multicolumn{2}{c}{$\St$-RKM variants} & \multirow{2}{*}{Liu2020} & \multirow{2}{*}{PCA} & \multirow{2}{*}{VAE} & \multirow{2}{*}{GAN}\\ \cmidrule{2-3}
     &  $E_{\text{FullEnergy}}$ & $E_{\text{kPCAError}}$ & && \\ \midrule
FPR95\small{($\uparrow$)}& {92.68}  (0.1) & 92.76  (0.2)  & 92.54  (0.2)  & 95.13  & 80.11  (0.2) & \textbf{95.07}  (0.3) \\
AUROC\small{($\uparrow$)}  & 54.08  (0.1) & 52.72  (0.1)  & 51.23  (0.1)  & 49.66  & \textbf{65.81}  (0.2) & 50.11  (0.3) \\
AUPR\small{($\uparrow$)}  &  87.36  (0.02) & 86.54  (0.04)  & 85.91  (0.1)  & 85.57  & \textbf{90.64}  (0.1) & 85.82  (0.1) \\
\bottomrule
\end{tabular}
\end{table*}

\begin{figure*}[h]
	\centering
	\includegraphics[width=0.8\textwidth]{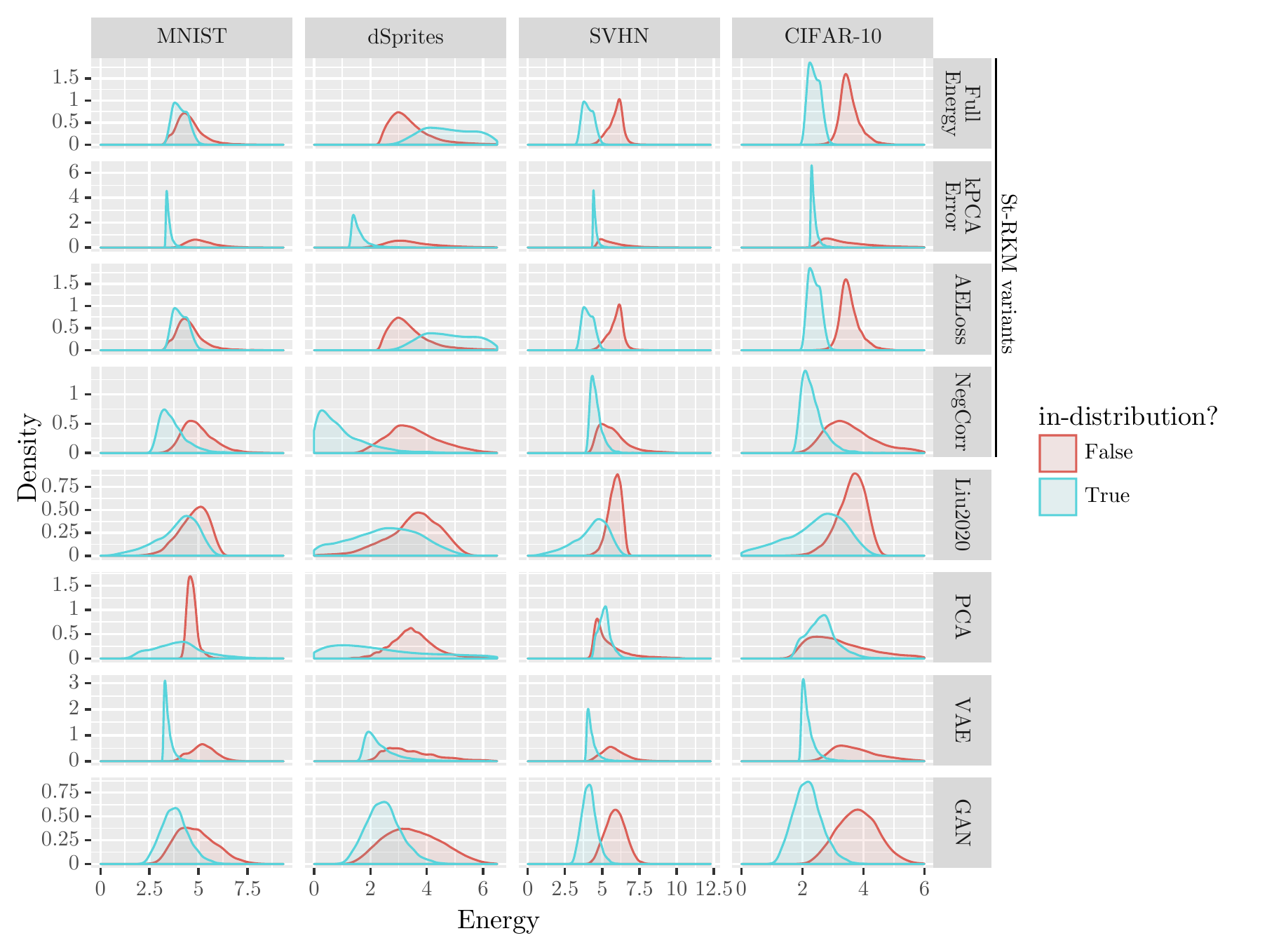}
	\caption{Distribution of the energy scores for the proposed energy functions (first four rows) based on the St-RKM's energy, for the state-of-the-art energy-based OOD detector without fine-tuning proposed in \cite{liu2020a}, for a PCA baseline and for the generative-based methods VAE and GAN. Note that, for MNIST and dSprites, the in- and out- distributions overlap for the \textit{AELoss} definition, while they are separated for CIFAR-10 and SVHN datasets. At the same time, the distributions for the $E_{\text{FullEnergy}}$ definition follow the distributions for the \textit{AELoss} definition very closely, showing that in those datasets the autoencoder loss term has a dominant weight in the energy function of the $E_{\text{FullEnergy}}$ definition.}
\end{figure*}

Next, we further analyze the observations in Fig. \ref{fig:sanitycheck}, and report the FPR95, AUROC and AUPR  scores in Table \ref{tab:sanitycheck}. 
It is important to note that in this experiment, higher values of FPR95 are better since the test points used for evaluation, which are assumed to be OOD, are actually sampled from the in-distribution. In other words, the samples from the test set of Fashion-MNIST are considered OOD in the evaluation. Accordingly, an FPR of 90\% means that 90 out of 100 samples from the test set of Fashion-MNIST were classified as in-distribution. The values in the table approach 95\% because we set the threshold such that TPR is 95\% and in this experiment TPR = FPR. 
Values of AUROC are around 50\% because both positive and negative instances correspond to in-distribution samples. For this reason and because the ratio of positive to negative instances is 6, values of AUPR are around $\frac{6}{7}$.

\begin{figure*}[t!]
	\centering
	\includegraphics[width=\textwidth]{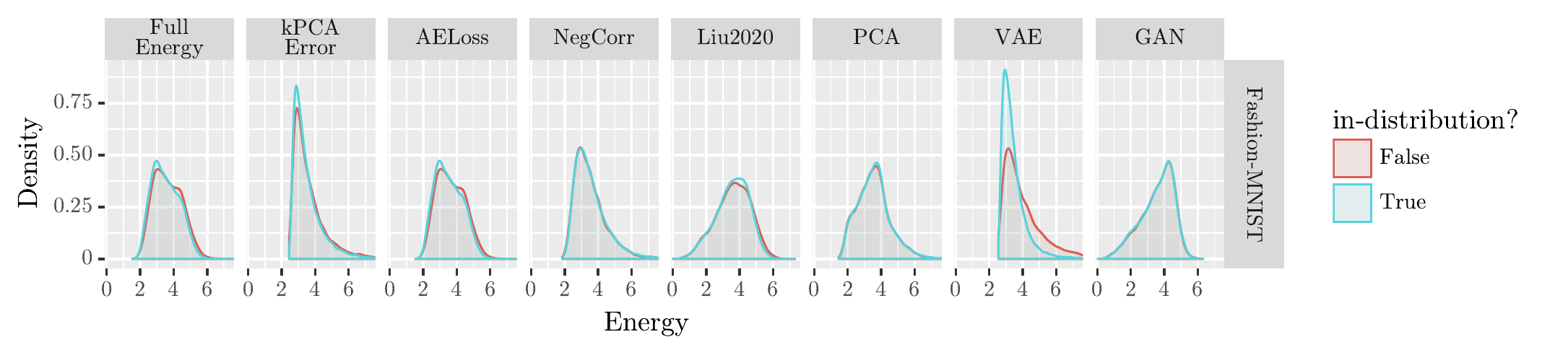}
	\caption{Distribution of the energy scores when the in-distribution coincides with the out-distribution. In this experiment, the $\Dtrainin$ is the training set of Fashion-MNIST ($N_{\text{train}} = 60,000$) and $\Dtestout$ is the test set ($N_{\text{test}} = 10,000$) of the same dataset.}
\end{figure*}

\end{document}

%% file: drawing_ood_cifar.pdf_tex
\newcommand{\highlight}[1]{\colorbox{yellow}{$\displaystyle #1$}}
\begingroup%
  \makeatletter%
  \providecommand\color[2][]{%
    \errmessage{(Inkscape) Color is used for the text in Inkscape, but the package 'color.sty' is not loaded}%
    \renewcommand\color[2][]{}%
  }%
  \providecommand\transparent[1]{%
    \errmessage{(Inkscape) Transparency is used (non-zero) for the text in Inkscape, but the package 'transparent.sty' is not loaded}%
    \renewcommand\transparent[1]{}%
  }%
  \providecommand\rotatebox[2]{#2}%
  \newcommand*\fsize{\dimexpr\f@size pt\relax}%
  \newcommand*\lineheight[1]{\fontsize{\fsize}{#1\fsize}\selectfont}%
  \ifx\svgwidth\undefined%
    \setlength{\unitlength}{400.12220007bp}%
    \ifx\svgscale\undefined%
      \relax%
    \else%
      \setlength{\unitlength}{\unitlength * \real{\svgscale}}%
    \fi%
  \else%
    \setlength{\unitlength}{\svgwidth}%
  \fi%
  \global\let\svgwidth\undefined%
  \global\let\svgscale\undefined%
  \makeatother%
  \setlength{\fboxsep}{0.1em}
  \newcommand{\cfbox}[2]{%
    \colorlet{currentcolor}{.}%
    {\color{#1}%
    \fbox{\color{currentcolor}#2}}%
}
  \begin{picture}(1,0.42222445)%
    \lineheight{1}%
    \setlength\tabcolsep{0pt}%
    \put(0,0){\includegraphics[width=\unitlength,page=1]{drawing_ood_cifar.pdf}}%
    \put(0.08679879,0.3050959){\color[rgb]{0,0,0}\makebox(0,0)[lt]{\lineheight{1.25}\smash{\begin{tabular}[t]{l}$x_1$\end{tabular}}}}%
    \put(0.20767984,0.24861355){\color[rgb]{0,0,0}\makebox(0,0)[lt]{\lineheight{1.25}\smash{\begin{tabular}[t]{l}$x_n$\end{tabular}}}}%
    \put(0.066809,0.21226859){\color[rgb]{0,0,0}\makebox(0,0)[lt]{\lineheight{1.25}\smash{\begin{tabular}[t]{l}$x_3$\end{tabular}}}}%
    \put(0.10520733,0.23251875){\color[rgb]{0,0,0}\makebox(0,0)[lt]{\lineheight{1.25}\smash{\begin{tabular}[t]{l}$x_2$\end{tabular}}}}%
    \put(0.19830168,0.29104517){\color[rgb]{0,0,0}\makebox(0,0)[lt]{\lineheight{1.25}\smash{\begin{tabular}[t]{l}   \cfbox{green}{$x$}\end{tabular}}}}%
    
    \put(0.19830168,0.17283015){\color[rgb]{0,0,0}\makebox(0,0)[lt]{\lineheight{1.25}\smash{\begin{tabular}[t]{l}\cfbox{green}{$\tilde{x}$}\end{tabular}}}}%
    \put(0.80607605,0.33220415){\color[rgb]{0,0,0}\makebox(0,0)[lt]{\lineheight{1.25}\smash{\begin{tabular}[t]{l}$\phi_{\theta}(x_1)$\end{tabular}}}}%
    \put(0.80046479,0.22719039){\color[rgb]{0,0,0}\makebox(0,0)[lt]{\lineheight{1.25}\smash{\begin{tabular}[t]{l}$\phi_{\theta}(x_2)$\end{tabular}}}}%
    \put(0.66814321,0.08477879){\color[rgb]{0,0,0}\makebox(0,0)[lt]{\lineheight{1.25}\smash{\begin{tabular}[t]{l}$\phi_{\theta}(x_3)$\end{tabular}}}}%
    \put(0.70308041,0.04382122){\color[rgb]{0,0,0}\makebox(0,0)[lt]{\lineheight{1.25}\smash{\begin{tabular}[t]{l}$\phi_{\theta}(x_n)$\end{tabular}}}}%
    \put(0.67995709,0.32878296){\color[rgb]{0,0,0}\makebox(0,0)[lt]{\lineheight{1.25}\smash{\begin{tabular}[t]{l}$\phi_{\theta}(x)$\end{tabular}}}}%
    \put(0.73564966,0.14991112){\color[rgb]{0,0,0}\makebox(0,0)[lt]{\lineheight{1.25}\smash{\begin{tabular}[t]{l}$h$\end{tabular}}}}%
    \put(0.07324542,0.38314404){\color[rgb]{0,0,0}\makebox(0,0)[lt]{\lineheight{1.25}\smash{\begin{tabular}[t]{l}Input space\end{tabular}}}}%
    \put(0.05224893,0.13530669){\color[rgb]{0,0,0}\makebox(0,0)[lt]{\lineheight{1.25}\smash{\begin{tabular}[t]{l}$\mathbb{R}^{D}$\end{tabular}}}}%
    \put(0.54195609,0.02887811){\color[rgb]{0,0,0}\makebox(0,0)[lt]{\lineheight{1.25}\smash{\begin{tabular}[t]{l}$\mathbb{R}^{\ell}$\end{tabular}}}}%
    \put(0.40407177,0.30241579){\color[rgb]{0,0,0}\makebox(0,0)[lt]{\lineheight{1.25}\smash{\begin{tabular}[t]{l}$\phi_{\theta}(\cdot)$\end{tabular}}}}%
    \put(0.40407177,0.150426){\color[rgb]{0,0,0}\makebox(0,0)[lt]{\lineheight{1.25}\smash{\begin{tabular}[t]{l}$\psi_{\xi}(\cdot)$\end{tabular}}}}%
    \put(0.88844413,0.15189097){\color[rgb]{0,0,0}\makebox(0,0)[lt]{\lineheight{1.25}\smash{\begin{tabular}[t]{l}range($U$)\end{tabular}}}}%
    \put(0.65369465,0.37764231){\color[rgb]{0,0,0}\makebox(0,0)[lt]{\lineheight{1.25}\smash{\begin{tabular}[t]{l}Latent space\end{tabular}}}}%
    
    \put(0.64667728,0.24061162){\color[rgb]{0,0,0}\makebox(0,0)[lt]{\lineheight{1.25}\smash{\begin{tabular}[t]{l}$\mathbb{P}_{U}$\end{tabular}}}}%
    \put(0.8990972,0.1267738){\color[rgb]{0,0,0}\makebox(0,0)[lt]{\lineheight{1.25}\smash{\begin{tabular}[t]{l}$u_1$\end{tabular}}}}%
    \put(0.60056087,0.28947226){\color[rgb]{0,0,0}\makebox(0,0)[lt]{\lineheight{1.25}\smash{\begin{tabular}[t]{l}$u_2$\end{tabular}}}}%
  \end{picture}%
\endgroup%

%% file: main.bbl
\begin{thebibliography}{10}
\providecommand{\url}[1]{#1}
\csname url@samestyle\endcsname
\providecommand{\newblock}{\relax}
\providecommand{\bibinfo}[2]{#2}
\providecommand{\BIBentrySTDinterwordspacing}{\spaceskip=0pt\relax}
\providecommand{\BIBentryALTinterwordstretchfactor}{4}
\providecommand{\BIBentryALTinterwordspacing}{\spaceskip=\fontdimen2\font plus
\BIBentryALTinterwordstretchfactor\fontdimen3\font minus
  \fontdimen4\font\relax}
\providecommand{\BIBforeignlanguage}[2]{{%
\expandafter\ifx\csname l@#1\endcsname\relax
\typeout{** WARNING: IEEEtran.bst: No hyphenation pattern has been}%
\typeout{** loaded for the language `#1'. Using the pattern for}%
\typeout{** the default language instead.}%
\else
\language=\csname l@#1\endcsname
\fi
#2}}
\providecommand{\BIBdecl}{\relax}
\BIBdecl

\bibitem{goodfellow2015}
I.~J. Goodfellow, J.~Shlens, and C.~Szegedy, ``Explaining and harnessing
  adversarial examples,'' in \emph{The 3rd {{International Conference}} on
  {{Learning Representations}}}, 2015.

\bibitem{hendrycks2017}
D.~Hendrycks and K.~Gimpel, ``A {{Baseline}} for {{Detecting Misclassified}}
  and {{Out}}-of-{{Distribution Examples}} in {{Neural Networks}},'' in
  \emph{The 5th {{International Conference}} on {{Learning Representations}}},
  2017.

\bibitem{rousseeuw2011}
P.~J. Rousseeuw and M.~Hubert, ``\BIBforeignlanguage{en}{Robust statistics for
  outlier detection},'' \emph{\BIBforeignlanguage{en}{WIREs Data Mining and
  Knowledge Discovery}}, vol.~1, no.~1, pp. 73--79, 2011.

\bibitem{hodge2004}
V.~J. Hodge and J.~Austin, ``\BIBforeignlanguage{en}{A {{Survey}} of {{Outlier
  Detection Methodologies}}},'' \emph{\BIBforeignlanguage{en}{Artificial
  Intelligence Review}}, vol.~22, no.~2, pp. 85--126, 2004.

\bibitem{robustrkm}
A.~Pandey, J.~Schreurs, and J.~A.~K. Suykens, ``Robust {{Generative Restricted
  Kernel Machines}} using {{Weighted Conjugate Feature Duality}},'' in
  \emph{International {{Conference}} on {{Machine Learning}}, {{Optimization}},
  and {{Data Science}}}, 2020.

\bibitem{strkm}
A.~Pandey, M.~Fanuel, J.~Schreurs, and J.~A.~K. Suykens, ``Disentangled
  {{Representation Learning}} and {{Generation}} with {{Manifold
  Optimization}},'' \emph{arXiv:2006.07046 [cs, stat]}, 2020.

\bibitem{liu2020a}
W.~Liu, X.~Wang, J.~D. Owens, and Y.~Li, ``Energy-based {{Out}}-of-distribution
  {{Detection}},'' in \emph{Proceedings of the 34th {{International
  Conference}} on {{Neural Information Processing Systems}}}, 2020.

\bibitem{pca}
I.~T. Jolliffe, \emph{Principal Components Analysis}.\hskip 1em plus 0.5em
  minus 0.4em\relax {Springer}, 1986.

\bibitem{pimentel2014}
M.~A.~F. Pimentel, D.~A. Clifton, L.~Clifton, and L.~Tarassenko,
  ``\BIBforeignlanguage{en}{A review of novelty detection},''
  \emph{\BIBforeignlanguage{en}{Signal Processing}}, vol.~99, pp. 215--249,
  2014.

\bibitem{liang2018}
S.~Liang, Y.~Li, and R.~Srikant, ``Enhancing {{The Reliability}} of
  {{Out}}-of-distribution {{Image Detection}} in {{Neural Networks}},'' in
  \emph{The 6th {{International Conference}} on {{Learning Representations}}},
  2018.

\bibitem{lee2018}
K.~Lee, K.~Lee, H.~Lee, and J.~Shin, ``A simple unified framework for detecting
  out-of-distribution samples and adversarial attacks,'' in \emph{Proceedings
  of the 32nd {{International Conference}} on {{Neural Information Processing
  Systems}}}, 2018.

\bibitem{hendrycks2019}
D.~Hendrycks, M.~Mazeika, and T.~Dietterich, ``Deep {{Anomaly Detection}} with
  {{Outlier Exposure}},'' in \emph{The 7th {{International Conference}} on
  {{Learning Representations}}}, 2019.

\bibitem{skvara2018}
V.~{\v S}kv{\'a}ra, T.~Pevn{\'y}, and V.~{\v S}m{\'i}dl, ``Are generative deep
  models for novelty detection truly better?'' in \emph{{{The 24th ACM SIGKDD
  International Conference on Knowledge Discovery and Data Mining: OOD
  Workshop}}}, 2018.

\bibitem{vae}
D.~P. Kingma and M.~Welling, ``Auto-encoding variational bayes,'' in \emph{The
  2nd {{International Conference}} on {{Learning Representations}}}, 2014.

\bibitem{gan}
I.~Goodfellow, J.~{Pouget-Abadie}, M.~Mirza, B.~Xu, D.~{Warde-Farley},
  S.~Ozair, A.~Courville, and Y.~Bengio, ``Generative adversarial nets,'' in
  \emph{Proceedings of the 28th {{International Conference}} on {{Neural
  Information Processing Systems}}}, 2014.

\bibitem{nalisnick2019}
E.~Nalisnick, A.~Matsukawa, Y.~W. Teh, D.~Gorur, and B.~Lakshminarayanan, ``Do
  {{Deep Generative Models Know What They Don}}'t {{Know}}?'' in \emph{The 7th
  {{International Conference}} on {{Learning Representations}}}, 2019.

\bibitem{lecun2006}
Y.~LeCun, S.~Chopra, R.~Hadsell, M.~Ranzato, and F.~Huang, ``A tutorial on
  energy-based learning,'' in \emph{Predicting Structured Data}, G.~BakIr,
  T.~Hofmann, B.~Sch{\"o}lkopf, A.~J. Smola, B.~Taskar, and S.~N. Vishwanathan,
  Eds.\hskip 1em plus 0.5em minus 0.4em\relax {MIT Press}, 2006, pp. 191--246.

\bibitem{grathwohl2020}
W.~Grathwohl, K.-C. Wang, J.-H. Jacobsen, D.~Duvenaud, M.~Norouzi, and
  K.~Swersky, ``Your classifier is secretly an energy based model and you
  should treat it like one,'' in \emph{The 8th {{International Conference}} on
  {{Learning Representations}}}, 2020.

\bibitem{drkm}
J.~A.~K. Suykens, ``Deep {{Restricted Kernel Machines Using Conjugate Feature
  Duality}},'' \emph{Neural Computation}, vol.~29, no.~8, pp. 2123--2163, 2017.

\bibitem{KingmaAdam}
D.~P. Kingma and J.~L. Ba, ``Adam: {{A}} method for stochastic optimization,''
  in \emph{The 3rd {{International Conference}} on {{Learning
  Representations}}}, 2015.

\bibitem{Li2020Efficient}
J.~Li, F.~Li, and S.~Todorovic, ``Efficient {{Riemannian Optimization}} on the
  {{Stiefel Manifold}} via the {{Cayley Transform}},'' in \emph{The 8th
  {{International Conference}} on {{Learning Representations}}}, 2019.

\bibitem{inman1989}
H.~F. Inman and E.~L.~B. Jr, ``The overlapping coefficient as a measure of
  agreement between probability distributions and point estimation of the
  overlap of two normal densities,'' \emph{Communications in Statistics -
  Theory and Methods}, vol.~18, no.~10, pp. 3851--3874, 1989.

\bibitem{pastore2019}
M.~Pastore and A.~Calcagn{\`i}, ``\BIBforeignlanguage{English}{Measuring
  {{Distribution Similarities Between Samples}}: {{A Distribution}}-{{Free
  Overlapping Index}}},'' \emph{\BIBforeignlanguage{English}{Frontiers in
  Psychology}}, vol.~10, 2019.

\bibitem{fashionmnist}
H.~Xiao, K.~Rasul, and R.~Vollgraf, ``Fashion-{{MNIST}}: A {{Novel Image
  Dataset}} for {{Benchmarking Machine Learning Algorithms}},''
  \emph{arXiv:1708.07747 [cs, stat]}, 2017.

\bibitem{mnist}
Y.~LeCun, C.~Cortes, and C.~Burges, ``{{MNIST}} handwritten digit database,''
  http://yann.lecun.com/exdb/mnist, 2010.

\bibitem{betavae}
I.~Higgins, L.~Matthey, A.~Pal, C.~Burgess, X.~Glorot, M.~Botvinick,
  S.~Mohamed, and A.~Lerchner, ``Beta-{{VAE}}: {{Learning}} basic visual
  concepts with a constrained variational framework,'' in \emph{The 5th
  {{International Conference}} on {{Learning Representations}}}, 2017.

\bibitem{svhn}
Y.~Netzer, T.~Wang, A.~Coates, A.~Bissacco, B.~Wu, and A.~Y. Ng, ``Reading
  digits in natural images with unsupervised feature learning,'' in \emph{The
  25th {{International Conference}} on {{Neural Information Processing
  Systems}}. {{Workshop}} on {{Deep Learning}} and {{Unsupervised Feature
  Learning}}}, 2011.

\bibitem{cifar10}
A.~Krizhevsky, ``Learning {{Multiple Layers}} of {{Features}} from {{Tiny
  Images}},''
  \url{https://www.cs.toronto.edu/~kriz/learning-features-2009-TR.pdf}, 2009.

\bibitem{fabius2015}
O.~Fabius, J.~R. {van Amersfoort}, and D.~P. Kingma, ``Variational {{Recurrent
  Auto}}-{{Encoders}},'' in \emph{Workshop at {{The}} 3rd {{International
  Conference}} on {{Learning Representations}}}, 2015.

\bibitem{ecg5000}
Y.~Chen, E.~Keogh, B.~Hu, N.~Begum, A.~Bagnall, A.~Mueen, and G.~Batista, ``The
  {{UCR}} time series classification archive,''
  \url{https://www.cs.ucr.edu/~eamonn/time_series_data/}, 2015.

\end{thebibliography}
